\documentclass[11pt]{article}
\PassOptionsToPackage{hyperfootnotes=false}{hyperref}
\usepackage[preprint]{acl}
\usepackage{times}
\usepackage{latexsym}
\usepackage[T1]{fontenc}
\usepackage[utf8]{inputenc}
\usepackage{microtype}
\usepackage{inconsolata}
\usepackage{graphicx}
\usepackage{booktabs}
\usepackage{tabularx}
\usepackage{multirow}
\usepackage{placeins}
\usepackage{amsmath,amssymb}
\usepackage{url}
\usepackage{enumitem}
\newcommand{\method}{\textsc{SEAF}}
\newcommand{\vcore}{A_0}
\newcommand{\tcore}{T_0}

\usepackage{iftex}
\usepackage{tcolorbox}
\tcbuselibrary{skins}
\usepackage{fvextra}
\ifPDFTeX
  \usepackage{CJKutf8}
\fi

\newtcolorbox{promptpanel}[1]{%
  enhanced,
  colback=blue!3!white,
  colframe=blue!40!black,
  boxrule=0.5pt,
  arc=1.5mm,
  left=2mm,right=2mm,top=2mm,bottom=2mm,
  before skip=7pt,after skip=4pt,
  fonttitle=\sffamily\bfseries\small,
  coltitle=blue!35!black,
  title={#1},
  attach boxed title to top center={yshift=-1.5mm},
  boxed title style={%
    colback=blue!10!white,
    colframe=blue!40!black,
    boxrule=0.5pt,arc=1mm,
    left=2mm,right=2mm,top=1mm,bottom=1mm
  }
}

\newsavebox{\seafpromptwordone}
\newsavebox{\seafpromptwordtwo}
\newsavebox{\seafpromptwordthree}
\newsavebox{\seafpromptwordfour}
\newsavebox{\seafpromptwordfive}
\newcommand{\seafpromptwordtext}[1]{%
  \ifPDFTeX
    \begin{CJK*}{UTF8}{gbsn}#1\end{CJK*}%
  \else
    #1%
  \fi
}
\AtBeginDocument{%
  \sbox{\seafpromptwordone}{\small\ttfamily\seafpromptwordtext{讽刺}}%
  \sbox{\seafpromptwordtwo}{\small\ttfamily\seafpromptwordtext{反讽}}%
  \sbox{\seafpromptwordthree}{\small\ttfamily\seafpromptwordtext{讥讽}}%
  \sbox{\seafpromptwordfour}{\small\ttfamily\seafpromptwordtext{嘲讽}}%
  \sbox{\seafpromptwordfive}{\small\ttfamily\seafpromptwordtext{反语}}%
}
\newcommand{\seafpromptterma}{\usebox{\seafpromptwordone}}
\newcommand{\seafprompttermb}{\usebox{\seafpromptwordtwo}}
\newcommand{\seafprompttermc}{\usebox{\seafpromptwordthree}}
\newcommand{\seafprompttermd}{\usebox{\seafpromptwordfour}}
\newcommand{\seafpromptterme}{\usebox{\seafpromptwordfive}}

\title{SEAF: Selective Semantic Enrichment and Visual Evidence Fusion\\for Multimodal Sarcasm Detection}
\author{
  Faxian Wan, Xiaocui Yang\thanks{Corresponding author.}, Yifan Cao, \\
  \textbf{Shi Feng, Daling Wang, Yifei Zhang} \\
  \normalfont School of Computer Science and Engineering, Northeastern University \\
  Shenyang, 110819, China \\
  \texttt{wanfaxian@mails.neu.edu.cn}, \texttt{yangxiaocui@cse.neu.edu.cn} \\
  \texttt{caoyifan@mails.neu.edu.cn}, \texttt{fengshi@cse.neu.edu.cn} \\
  \texttt{wangdaling@cse.neu.edu.cn}, \texttt{zhangyifei@cse.neu.edu.cn}
}
\hypersetup{
  pdftitle={SEAF: Selective Semantic Enrichment and Visual Evidence Fusion for Multimodal Sarcasm Detection},
  pdfauthor={Faxian Wan, Xiaocui Yang, Yifan Cao, Shi Feng, Daling Wang, Yifei Zhang},
  pdfsubject={Multimodal sarcasm detection},
  pdfkeywords={multimodal sarcasm detection, semantic enrichment, visual evidence fusion}
}
\begin{document}
\maketitle

\begin{abstract}
Multimodal sarcasm detection requires textual, visual, and contextual evidence, particularly when multiple images provide complementary cues. 
Multimodal large language models (MLLMs) can generate auxiliary information from a post's text and images, but their outputs may mix useful background knowledge with redundant visual descriptions. 
We propose \textbf{SEAF}, a dual-branch framework for selective semantic enrichment and visual evidence fusion. SEAF assigns generated fields to distinct roles: entity background enriches semantic input, while image descriptions supply alignment statistics. 
The semantic branch combines the original post text, background from a frozen MLLM, and optical character recognition (OCR) text. The aligned visual branch combines global and local visual features with statistics of cross-modal and inter-image similarity. 
Text-Anchored Smoothed Aggregation (TASA) weights images by their similarity to the original post text while preserving contributions from images with low literal similarity. 
Summing the branch classification scores integrates contextual semantics with visual evidence. 
Experiments on the multi-image MMSD3.0 benchmark show that SEAF outperforms all compared methods.

\end{abstract}

\section{Introduction}

Multimodal sarcasm detection (MSD) determines whether a post is sarcastic by jointly interpreting its text and associated images. 
MMSD2.0 considers a single-image setting \citep{qin2023mmsd2}, whereas MMSD3.0 extends the task to multi-image posts \citep{Zhao2026}. 
\begin{figure}[t]
\centering
\includegraphics[width=\columnwidth]{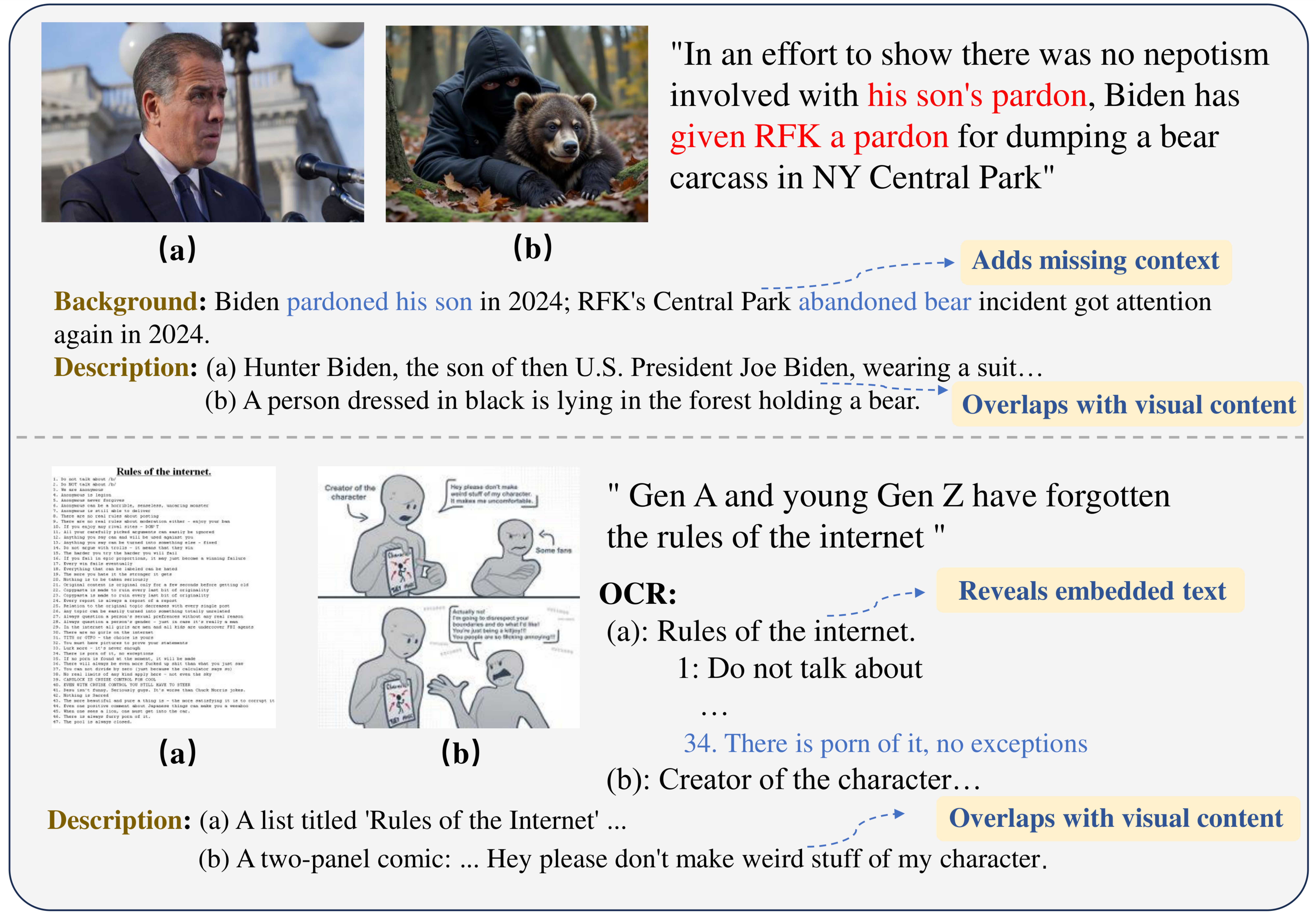}
\caption{Two sarcastic MMSD3.0 posts: background adds missing context (top), OCR reveals image text (bottom), and generated descriptions overlap with visible content. Labels (a) and (b) denote the two images within each post.}
\label{fig:motivation}
\end{figure}

Figure~\ref{fig:motivation} illustrates why different evidence sources require different treatment.
In the upper example, background connects Biden's pardon of his son with RFK's bear incident, clarifying how the invented pardon mocks the denial of nepotism.
Describing the people and bear alone leaves this contextual link unexplained.
In the lower example, optical character recognition (OCR) recovers the rules and comic dialogue across the two images, making the post text's reference to internet rules interpretable.
A generic description of a rule list and comic can miss the wording needed to connect them.
Meanwhile, descriptions generated by multimodal large language models (MLLMs) can repeat visible content already represented by image features.
These examples call for supplementing missing context and embedded text while retaining and aggregating complementary visual evidence.

Existing approaches strengthen cross-modal interactions through attention, graph-based modeling, and gating \citep{pan2020incongruity,wen2023dip,zhu2024dglf,zhang2026ghin}, or leverage MLLMs to generate auxiliary information \citep{line2024cofipara,tang2024retrieval,jana2025midre}. Incorporating all generated fields into a single semantic sequence, however, can blur their distinct roles and duplicate content already captured by visual features. The central design question is therefore \textit{which information should enrich semantic interpretation and how it should complement retained visual evidence}.

To address this question, we propose \textbf{\method{}}, a dual-branch architecture that assigns auxiliary information to complementary roles through a predefined field-allocation rule. The \textit{Semantic Evidence Branch} enriches the original post text with OCR text and selected entity background information. The \textit{Aligned Visual Evidence Branch} retains global and local visual features and uses generated image descriptions to compute alignment statistics. These descriptions support visual alignment without entering the semantic sequence.
For multi-image posts, we introduce \textit{Text-Anchored Smoothed Aggregation} (TASA), which weights image contributions according to their similarity to the original post text. Rather than suppressing images with low literal similarity, TASA smooths these weights to maintain a nonzero contribution from every valid image, allowing potentially sarcasm-relevant cues to remain available for prediction. Finally, \method{} sums the semantic and visual classifiers' scores to combine contextual interpretation with retained visual evidence for sarcasm prediction.
Our contributions are threefold:
\begin{itemize}[leftmargin=*, itemsep=1pt, topsep=1pt, parsep=0pt]
\item We propose \method{}, a dual-branch architecture for selective semantic enrichment and visual evidence fusion. A predefined field-allocation rule assigns entity background to semantic input and image descriptions to alignment statistics, while classification-score fusion combines the branches.
\item We introduce Text-Anchored Smoothed Aggregation (TASA), which weights images by their similarity to the original post text, retaining potentially sarcasm-relevant evidence with low literal similarity.
\item Experiments show the best overall MMSD3.0 performance among compared methods and assess single-image applicability and cross-dataset transfer. Component ablations and input controls establish the value of selective enrichment and retained visual evidence.
\end{itemize}

\begin{figure*}[t]
\centering
\includegraphics[width=\textwidth]{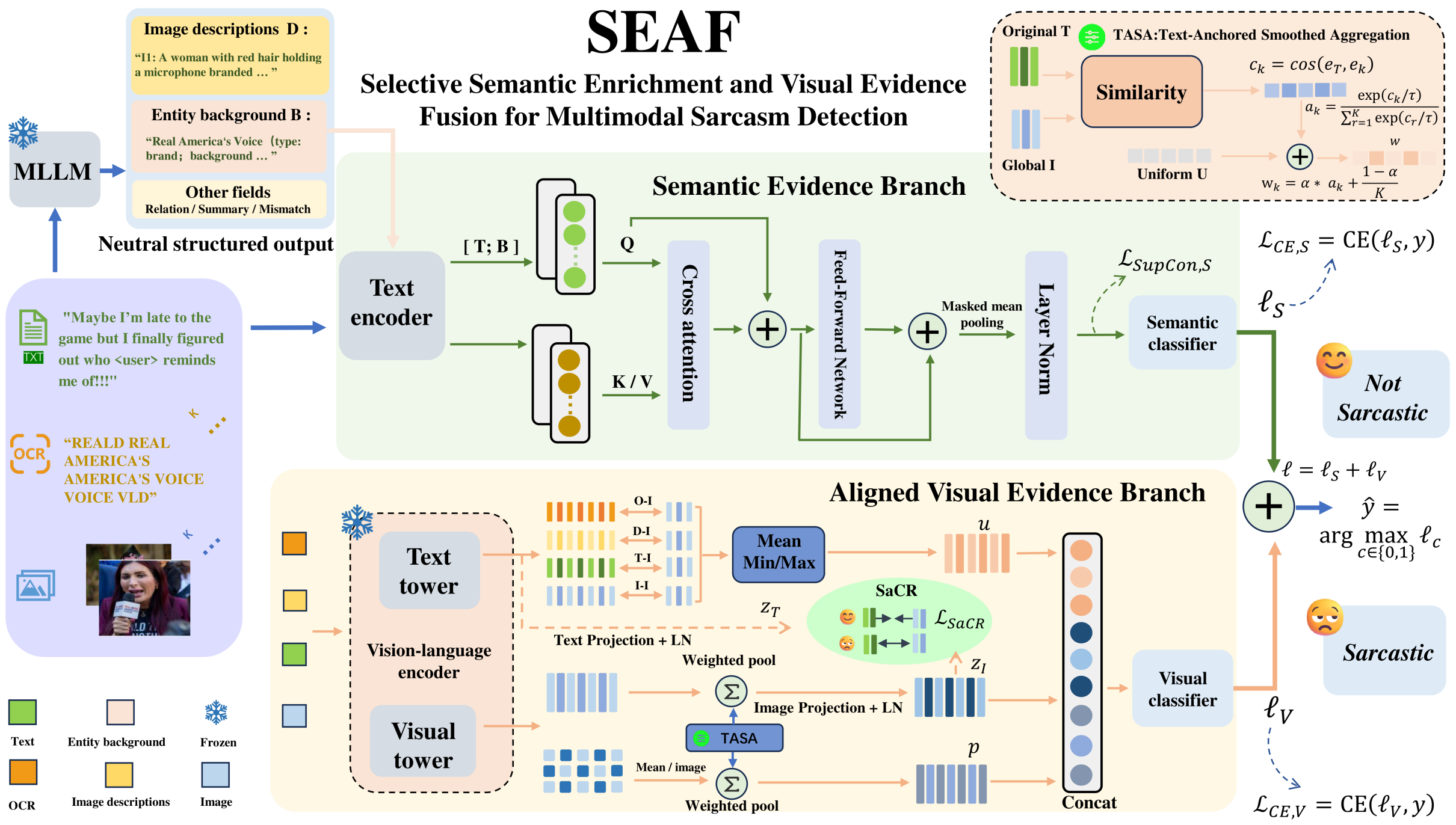}
\caption{SEAF architecture. The frozen multimodal large language model (MLLM) supplies neutral entity background and image descriptions for semantic enrichment and visual alignment, respectively. 
The two branches fuse classification scores for sarcasm prediction.}
\label{fig:overview}
\end{figure*}

\section{Related Work}

\paragraph{Multimodal Sarcasm Detection.}

MMSD provides an image--text benchmark for multimodal sarcasm detection (MSD) \citep{cai2019mmsd}, while MMSD2.0 improves annotation reliability and introduces multi-view representations \citep{qin2023mmsd2}.
Subsequent methods explore complementary forms of multimodal incongruity \citep{li-etal-2025-ambiguity,guo-etal-2025-multi,zhang-etal-2025-incongruity}.
MMSD3.0 extends MSD to posts with two to four images, requiring inter-image semantic and affective reasoning \citep{Zhao2026}.
CIRM addresses these dependencies through dual-stage bridging and relevance-guided fusion, but lacks explicit external analyses or entity background, potentially limiting recognition of sarcasm requiring event-specific or cultural knowledge.
SEAF addresses this contextual gap by adding neutrally generated entity background to the post text in its semantic branch.
Additional context offers benefits but can introduce biases \citep{nimase-hong-2024-contexts}; other approaches use generated reasoning as auxiliary evidence \citep{qiu-etal-2025-detecting,yang-etal-2025-sarcasm,GRASP2026,goel-etal-2025-target}.
However, using complete generated explanations as semantic input may mix supplementary background with visual content already represented by image features.
SEAF limits this duplication by using image descriptions for alignment statistics alongside retained global and local visual features.
For multi-image aggregation, smoothed, text-anchored weighting preserves contributions from weakly aligned images that may still carry sarcasm-relevant cues.

\section{SEAF}
\label{sec:method}
Figure~\ref{fig:overview} summarizes SEAF. The Semantic Evidence Branch combines post text, neutral entity background, and OCR to recover contextual and embedded-text cues.
The Aligned Visual Evidence Branch combines global and local features with alignment statistics to represent visual content and its relations to textual evidence.
TASA pools images while preserving weakly aligned cues; Sarcasm-aware Contrastive Regularization (SaCR) guides label-dependent image--text alignment during training.
Summing the branch scores produces the final prediction.

\subsection{Task Formulation}

Each multimodal post contains a text component $T$ and $K$ valid images $\mathcal I=\{I_k\}_{k=1}^{K}$; padded image slots are excluded from $K$.
The task is to determine whether the post is sarcastic ($y=1$) or non-sarcastic ($y=0$).
For each image $I_k$, the OCR text is denoted by $O_k$.
During preprocessing for training and inference, the frozen MLLM generates neutral entity background $B$ for the post and per-image descriptions $D_k$:
\begin{equation}
(B,\mathcal D)=\mathcal G_{\mathrm{neutral}}(T,\mathcal I),
\label{eq:neutral_generation}
\end{equation}
where $\mathcal D=\{D_k\}_{k=1}^{K}$ and $\mathcal G_{\mathrm{neutral}}$ denotes neutral analysis generation and field extraction.

\subsection{Neutral Descriptions and Background}
\label{sec:neutral}

Sarcasm may depend on event-specific or cultural knowledge absent from the post text and images.
The frozen MLLM supplies entity background and literal image descriptions as auxiliary evidence.
The neutral prompt prohibits sarcasm judgments and inferred author intent (Figure~\ref{fig:neutral_generation_prompt}).
SEAF appends entity background $B$ to the post text to contextualize its entities and references.
Image descriptions $\mathcal D$ supply description--image alignment statistics; global and local visual features represent image content directly.
This allocation is fixed across posts: SEAF selects generated fields by their role, while OCR attention and image weighting adapt their contributions to each input.
Other generated fields, including summaries and mismatch analyses, are excluded from SEAF's inputs.
Appendix~\ref{app:background} compares alternative generated fields.

\subsection{Semantic Evidence Branch}
\label{sec:semantic}

This branch learns a contextual representation by modeling interactions between post text augmented with entity background and OCR text.
Post text and background guide attention to relevant OCR tokens, while residual fusion preserves their contextual representation.
The post text $T$ and entity background $B$ are concatenated to form $T_B$:
\begin{equation}
T_B=T\mathbin{\Vert}\texttt{[KNOW]}\mathbin{\Vert}B,
\end{equation}
where $\Vert$ denotes text concatenation, and \texttt{[KNOW]} is a textual delimiter marking the beginning of the supplementary background knowledge.
No background is appended when the field is empty. OCR from valid images is concatenated in input order to form $O$. 
A shared semantic encoder $E_{\mathrm S}$ encodes $T_B$ and $O$. A shared projection maps both sequences to the fusion dimension, and distinct modality embeddings identify their sources, yielding token-representation matrices $H_T$ and $H_O$.
This encoder models contextual token interactions; the frozen CLIP text tower below instead supplies embeddings for image--text comparison.

In multi-head attention ($\operatorname{MHA}$), tokens from post text and background supply queries and OCR tokens supply keys and values. Residual cross-attention and a feed-forward network ($\operatorname{FFN}$) produce intermediate states $H_1$ and $H_2$:
\begin{align}
H_1&=H_T+\operatorname{MHA}
 \bigl(\operatorname{LN}_1(H_T),H_O,H_O\bigr),\\
H_2&=H_1+\operatorname{FFN}
 \bigl(\operatorname{LN}_2(H_1)\bigr).
\label{eq:semantic_residual}
\end{align}

Here, $\operatorname{LN}$ denotes layer normalization over the feature dimension; subscripts 1 and 2 distinguish the normalizations before attention and the feed-forward network.

Attention masks invalid OCR tokens. Masked mean pooling over valid $T_B$ tokens, followed by layer normalization, produces
\begin{align}
h_{\mathrm S}&=\operatorname{LN}
 \bigl(\operatorname{Mean}_{\mathrm{mask}}(H_2)\bigr),\\
\ell_{\mathrm S}&=f_{\mathrm S}
 \bigl(\operatorname{Dropout}(h_{\mathrm S})\bigr),
\end{align}
where $h_{\mathrm S}$ summarizes the semantic sequence and $f_{\mathrm S}$ is a linear classifier producing the two class logits $\ell_{\mathrm S}\in\mathbb R^2$. Supervised contrastive training also acts on $h_{\mathrm S}$.

\subsection{Aligned Visual Evidence Branch}
\label{sec:aligned_visual}
This branch models the overall content of the image set and variations in its text--image and image--image relations.
It combines pooled features with alignment statistics to preserve relational cues that may be obscured by visual pooling alone.

Frozen CLIP text and visual towers provide a shared image--text embedding space.
The text tower encodes the original post text $T$, per-image OCR $O_k$, and descriptions $D_k$ as $e_T$, $e_{O_k}$, and $e_{D_k}$; the visual tower encodes each image $I_k$ as a global vector $e_k$.
These vectors are $\ell_2$-normalized, so their dot products measure cosine similarity.
The visual tower also supplies cached local features $P_k=\{p_{kj}\}_{j=1}^{J}$, where $p_{kj}$ represents region $j$ of image $k$ on a $7\times7$ grid ($J=49$).

\subsubsection{Text-Anchored Smoothed Aggregation}
\label{sec:tasa}
TASA aggregates multiple images into a visual summary guided by the original post text.
Images can provide different evidence: some establish the post's subject, while others introduce contrasts.
Uniform pooling assigns them equal weight, while weighting only by text--image similarity may suppress contrasting images.
We therefore mix similarity weights with a uniform component, adapting image contributions while preserving weakly aligned cues.
The weights are computed as:

\begin{align}
c_k&=\cos(e_T,e_k)=e_T^\top e_k,\nonumber \\
a_k&=\frac{\exp(c_k/\tau)}
 {\sum_{r=1}^{K}\exp(c_r/\tau)},\nonumber\\
w_k&=\alpha a_k+\frac{1-\alpha}{K}.
\label{eq:weights}
\end{align}

Here, $c_k$ is image $k$'s similarity to the post text, $a_k$ is its softmax weight, and $w_k$ is its final aggregation weight. Temperature $\tau$ controls concentration, and $\alpha$ controls the mixture with uniform weights.
For $0\le\alpha<1$ and $\tau>0$, each valid image has weight $w_k\ge(1-\alpha)/K>0$; padded slots receive zero weight.
TASA computes these weights from frozen embeddings without a trainable predictor. 
For $K=1$, $w_1=1$. The aggregation is invariant to a consistent permutation of the images and their associated features.

Text--image similarity guides evidence allocation rather than directly indicating sarcasm relevance.
Mean-pooled patch embeddings provide complementary information about local image regions, helping retain visual cues that global embeddings may underrepresent.
We use the same weights to aggregate global image embeddings and mean-pooled patch embeddings from each image:
\begin{align}
g&=\sum_{k=1}^{K}w_ke_k,\\
z_I&=\operatorname{LN}(W_Ig+b_I), \\
p&=\sum_{k=1}^{K}w_k
 \left(\frac{1}{J}\sum_{j=1}^{J}p_{kj}\right).
\end{align}
Here, $g$ is the pooled global embedding, $W_I$ and $b_I$ are a trainable projection matrix and bias, and $z_I$ is the projected global representation.
The local summary $p$ averages regions within each image before TASA pools across images.

\subsubsection{Alignment Statistics and Classification}
\label{sec:signals}

To retain relations that visual pooling may obscure, this module measures each image's similarity to the post text, OCR text, its description, and other images:
\begin{equation}
q_k=\begin{bmatrix}
e_T^\top e_k\\
\mathbb I[O_k\neq\varnothing]\,e_{O_k}^\top e_k\\
\mathbb I[D_k\neq\varnothing]\,e_{D_k}^\top e_k\\
\max_{r\neq k}e_k^\top e_r
\end{bmatrix}.
\end{equation}
The four coordinates of $q_k$ are text--image (T--I), OCR--image (O--I), description--image (D--I), and image--image (I--I) similarity (Figure~\ref{fig:overview}).
The indicator $\mathbb I[\cdot]$ equals one when its condition holds and zero otherwise; missing OCR or descriptions contribute zero.
The last coordinate is the maximum similarity to another valid image, set to zero when $K=1$. 
Coordinate-wise mean, minimum, and maximum over valid images capture overall alignment and extreme values:
\begin{equation}
u=\left[\operatorname{mean}_kq_k;
 \operatorname{min}_kq_k;
 \operatorname{max}_kq_k\right]\in\mathbb R^{12}.
\end{equation}
Here, $[\cdot;\cdot]$ denotes vector concatenation. The four relations yield 12 statistics independently of TASA weights.
The visual classifier $f_{\mathrm V}$ is a two-layer perceptron with Gaussian error linear unit (GELU) activation and dropout:
\begin{equation}
\ell_{\mathrm V}=f_{\mathrm V}([z_I;p;u]).
\end{equation}
Its logits $\ell_{\mathrm V}\in\mathbb R^2$ combine global content ($z_I$), local content ($p$), and alignment relations ($u$).

\subsection{Score Fusion and Training}

Adding the branch logits lets contextual semantics and visual evidence contribute directly to the final decision:
\begin{equation}
 \ell=\ell_{\mathrm S}+\ell_{\mathrm V};  
\hat y=\arg\max_{c\in\{0,1\}}\ell_c,   
\end{equation}
where $\ell_c$ is the summed logit (unnormalized score) for class $c$, and $\hat y$ is the predicted label. Separate cross-entropy supervision trains each branch to produce discriminative class scores:
\begin{equation}
\mathcal L_{\mathrm{CE,S}}=\operatorname{CE}(\ell_{\mathrm S},y),
\mathcal L_{\mathrm{CE,V}}=\operatorname{CE}(\ell_{\mathrm V},y).
\end{equation}
Here, $\operatorname{CE}$ denotes cross-entropy with label smoothing.
To complement cross-entropy supervision, we apply supervised contrastive learning \citep{khosla2020supcon} to $h_{\mathrm S}$ before classification dropout, with loss $\mathcal L_{\mathrm{SupCon,S}}$.
This loss compares semantic representations across training examples, 
pulling together those with the same sarcasm label and separating those with different labels to encourage a discriminative semantic representation space.

\paragraph{Sarcasm-aware Contrastive Regularization.}
SaCR regularizes text--image alignment within each post, complementing the semantic constraint across examples.
Separate trainable projections with layer normalization map the original post text embedding $e_T$ and TASA-pooled global image features to a common alignment space:
\begin{equation}
\begin{aligned}
z_T&=\operatorname{LN}(W_Te_T+b_T),\\
z_I&=\operatorname{LN}\left(W_I\sum_{k=1}^{K}w_ke_k+b_I\right).
\end{aligned}
\label{eq:sacr-projections}
\end{equation}
Here, $W_T$ and $b_T$ are the text projection matrix and bias; $z_T$ has the same dimension as the visual classifier's $z_I$.
With projected cosine similarity $s=\cos(z_T,z_I)$ and margin $m>0$, SaCR applies the objective of \citet{guoDGCF}:
\begin{equation}
\begin{aligned}
\mathcal L_{\mathrm{SaCR}}
={}&y\max(0,s+m)\\
&+(1-y)\max(0,m-s).
\end{aligned}
\label{eq:sacr-loss}
\end{equation}
This encourages $s\ge m$ for non-sarcastic and $s\le-m$ for sarcastic examples. 
These are task-specific constraints in a learned projection space, not assumptions that the text in every sarcastic post is literally dissimilar to its images. 
SaCR supplies training-only supervision; its projected text vector $z_T$ is not a classifier input. The total objective, averaged over training examples, is
\begin{equation}
\begin{aligned}
\mathcal L={}&\mathcal L_{\mathrm{CE,S}}
 +\mathcal L_{\mathrm{CE,V}}\\
&+\lambda_{\mathrm{SC}}\mathcal L_{\mathrm{SupCon,S}}
 +\mathcal L_{\mathrm{SaCR}}.
\end{aligned}
\label{eq:training-objective}
\end{equation}
Here, $\lambda_{\mathrm{SC}}\ge0$ weights semantic supervised contrastive learning; the other losses have unit weight. Fused scores receive no additional cross-entropy.
Task supervision optimizes trainable semantic layers, projections, residual fusion, and classifiers, while the MLLM and CLIP towers remain frozen.

\section{Experimental Setup}
\label{sec:setup}

\subsection{Datasets and Metrics}

MMSD3.0 is the primary benchmark, with two to four images per post (Figure~\ref{fig:motivation}); 
MMSD2.0 evaluates single-image applicability. Both use official splits (Table~\ref{tab:data}). 
SarcNet is used only for transfer to its complete official Chinese--English test set.

\begin{table}[!htb]
\resizebox{0.45\textwidth}{!}{
\centering\normalsize
\setlength{\tabcolsep}{3pt}
\renewcommand{\arraystretch}{1.0}
\begin{tabularx}{\columnwidth}{@{}>{\raggedright\arraybackslash}Xrrr@{}}
\toprule
\textbf{Dataset} & \textbf{Train} & \textbf{Dev} & \textbf{Test}\\
\midrule
\textbf{MMSD3.0} & 7,583 & 1,626 & 1,624\\
\textbf{MMSD2.0} & 19,816 & 2,410 & 2,409\\
\textbf{SarcNet} (transfer) & -- & -- & 666\\
\bottomrule
\end{tabularx}
}
\caption{Official splits. SarcNet training and validation data are not used.}
\label{tab:data}

\end{table}

\begin{table*}[!t]
\centering\normalsize
\setlength{\tabcolsep}{4pt}
\renewcommand{\arraystretch}{1.0}
\begin{tabularx}{\textwidth}{@{}l>{\raggedright\arraybackslash}Xrrrr@{}}
\toprule
\textbf{Modality} & \textbf{Method} & \textbf{Acc. (\%)} & \textbf{P (\%)} & \textbf{R (\%)} & \textbf{F1 (\%)}\\
\midrule
\multirow{3}{*}{Text only} & Bi-LSTM   & 74.78 & 73.18 & 70.31 & 71.12\\
 & BERT \cite{BERT} & 79.00 & 78.20 & 79.35 & 78.46\\
 & RoBERTa \cite{RoBERTa} & 79.99 & 79.68 & 81.15 & 79.67\\
\midrule
\multirow{2}{*}{Image only} & ResNet \cite{ResNet} & 60.26 & 45.86 & 53.92 & 49.57\\
 & ViT \cite{dosovitskiy2021vit} & 64.09 & 50.41 & 52.22 & 51.30\\
\midrule
\multirow{7}{*}{MLLM (zero-shot)} 
& GPT-5.4 \cite{openai2026gpt54} & 73.86 & 76.03 & 68.22 & 68.78\\
& GLM-4.6V \cite{glm2025vl} & 75.19 & 74.18 & 72.46 & 73.01\\
 & Doubao-seed-2.0-pro \cite{seed2026seed2} & 71.06 & 70.16 & 67.02 & 67.49\\
 & LLaVA-1.5-7B \cite{llava} & 61.12 & 60.07 & 60.82 & 59.84\\
 & Qwen3.7-Max \cite{alibabacloud2026qwen37max} & 72.27 & 72.49 & 67.45 & 67.95\\
 & Qwen3.6-35B-A3B \cite{qwen36_35b_a3b} & 72.39 & 72.61 & 67.61 & 68.13\\
 & Qwen3-VL-8B \cite{qwen2025vl} & 65.33 & 63.84 & 59.00 & 57.99\\
\midrule
\multirow{5}{*}{\shortstack[l]{Supervised \\ multimodal models}} & DIP \cite{wen2023dip} & 81.83 & 78.20 & 74.13 & 76.11\\
 & Multi-view CLIP \cite{qin2023mmsd2} & 81.96 & 81.84 & 81.96 & 81.96\\
 & MoBA \cite{MoBA} & 77.59 & 77.74 & 74.27 & 75.13\\
 & \citet{tang2024retrieval} & 82.20 & 81.79 & 80.36 & 80.91\\
 & CIRM$^\dagger$ \cite{Zhao2026} & 83.44 & 82.65 & 82.44 & 82.54\\
 \midrule
 \multirow{1}{*}{\textbf{Ours}} 
 & \textbf{\method{}} & \textbf{86.76} & \textbf{86.00} & \textbf{86.36} & \textbf{86.17}\\
 \bottomrule
\end{tabularx}
\caption{MMSD3.0 comparison (\%). Bold indicates the best result for each metric. $\dagger$ marks reproduced results. 
Evaluation protocols and metric conventions are described in Appendix~\ref{app:literature}.}
\label{tab:main}
\end{table*}

For SEAF, we report accuracy (Acc.), macro-precision ($P_M$), macro-recall ($R_M$), and macro-F1 ($F_M$), our primary metric. 
Transfer uses the area under the receiver operating characteristic curve (AUROC) for ranking, plus macro-F1, accuracy, and positive-class F1 under the source-trained decision rule. 
All metrics are percentages. Definitions appear in Appendix~\ref{app:evaluation}; comparison protocols appear in Appendix~\ref{app:literature}.
Matched ablations retain data, caches, and base training settings while changing designated inputs or components.

\subsection{Implementation}

We use DeBERTaV3 \citep{he2023debertav3} for semantic encoding, frozen CLIP \citep{radford2021clip} for aligned features, and Qwen3-VL-8B for neutral auxiliary generation.
\method{} results are means over three runs, with checkpoints selected by positive-class F1 on the validation set.
Detailed implementation settings, hyperparameters, and run aggregation rules are provided in Appendix~\ref{app:implementation}.

\subsection{Comparisons}

Table~\ref{tab:main} compares \method{} with text-only, image-only, supervised multimodal, and zero-shot MLLM methods on MMSD3.0.
Table~\ref{tab:mmsd2} evaluates its single-image applicability through comparisons with representative multimodal methods on MMSD2.0.
Section~\ref{sec:architecture_analysis} separately examines how input organization affects performance on both datasets.

\section{Experimental Results}
\label{sec:results}

\subsection{Main Results}

\method{} achieves a mean macro-F1 of 86.17 on MMSD3.0 (Table~\ref{tab:main}), exceeding CIRM by 3.63 points and obtaining the best results across all four reported metrics.
Accuracy, precision, and recall improve over CIRM by 3.32, 3.35, and 3.92 points, respectively, showing consistent gains across the reported measures.
CIRM explicitly models text--image and image--image dependencies, making it a strong baseline for this multi-image setting.
SEAF combines externally generated entity background with retained visual features and alignment statistics, providing contextual evidence beyond the observed post text and images.
Section~\ref{sec:ablation} further examines the contributions of these complementary information sources.

The strongest zero-shot MLLM achieves an F1 of 73.01, 13.16 points below \method{}.
This gap supports a division of labor in which the MLLM provides neutral contextual information and task-trained classifiers learn the sarcasm decision.
In SEAF, entity background enriches semantic interpretation, while image descriptions inform visual alignment statistics; the original visual features remain available to the classifier.

\begin{table}[!htb]
\centering
\begingroup
\small
\setlength{\tabcolsep}{1.25pt}
\renewcommand{\arraystretch}{1.3}
\begin{tabularx}{\columnwidth}{@{}>{\raggedright\arraybackslash}Xrrrr@{}}
\toprule
\textbf{Method} & \shortstack[r]{\textbf{Acc.}\\\textbf{(\%)}} & \shortstack[r]{\textbf{P}\\\textbf{(\%)}} & \shortstack[r]{\textbf{R}\\\textbf{(\%)}} & \shortstack[r]{\textbf{F1}\\\textbf{(\%)}}\\
\midrule
\multicolumn{5}{@{}l}{\textit{Zero-shot MLLM}}\\
GPT-5.4 \cite{openai2026gpt54} & 72.85 & 78.79 & 75.78 & 72.55 \\
\midrule
\multicolumn{5}{@{}l}{\textit{Supervised multimodal models}}\\
HFM \cite{cai2019mmsd} & 70.57 & 64.84 & 69.05 & 66.88\\
Att-BERT \cite{pan2020incongruity} & 80.03 & 76.28 & 77.82 & 77.04\\
HKE \cite{liu2022towards} & 76.50 & 73.48 & 71.07 & 72.25\\
DIP \cite{wen2023dip} & 80.96 & 78.02 & 77.56 & 77.79\\
Multi-view CLIP \cite{qin2023mmsd2} & 85.64 & 80.33 & 88.24 & 84.10\\
MoBA \cite{MoBA} & 85.22 & 79.82 & \textbf{88.29} & 84.11\\
ESAM \cite{yuan2025esam} & \textbf{85.87} & 83.12 & 86.05 & 84.56 \\
DGCF-B/32 \cite{guoDGCF} & 83.06 & 83.08 & 83.73 & 82.98\\
DGCF-L/336 \cite{guoDGCF} & 83.89 & 84.21 & 84.84 & 83.85\\
CIRM$^\dagger$ \cite{Zhao2026} & 81.11 & 81.08 & 81.69 & 81.01\\
\midrule
\textbf{\method{} (ours)} & 85.68 & \textbf{85.35} & 85.53 & \textbf{85.43}\\
\bottomrule
\end{tabularx}
\endgroup
\caption{MMSD2.0 results (\%). Bold indicates the best reported result in each column. $\dagger$ marks reproduced results.}
\label{tab:mmsd2}
\end{table}

\method{} achieves a mean macro-F1 of 85.43 and an accuracy of 85.68 on MMSD2.0 (Table~\ref{tab:mmsd2}), exceeding CIRM, DGCF-B/32, and DGCF-L/336 by 4.42, 2.45, and 1.58 macro-F1 points, respectively.
For a single image, TASA assigns unit weight, while semantic enrichment and visual alignment remain active.
Removing entity background or OCR lowers macro-F1 by 1.21 or 1.37 points, respectively (Table~\ref{tab:ablation_complete}), showing that contextual knowledge and image text remain useful in the single-image setting.

\subsection{Cross-Dataset Transfer}

\begin{table}[!htb]
\centering\normalsize
\setlength{\tabcolsep}{1.5pt}
\renewcommand{\arraystretch}{1.0}
\begin{tabularx}{\columnwidth}{@{}>{\raggedright\arraybackslash}Xrrrr@{}}
\toprule
\textbf{Model} & \textbf{AUROC(\%)} & \textbf{\boldmath$F_M$(\%)} & \textbf{Acc.(\%)} & \textbf{\boldmath$F_+$(\%)}\\
\midrule
\multicolumn{5}{l}{\textit{Trained on MMSD3.0}}\\
SEAF-S & 79.65 & \textbf{70.11} & \textbf{70.12} & \textbf{70.52}\\
\method{} & \textbf{84.29} & 67.25 & 67.42 & 64.94\\
\midrule
\multicolumn{5}{l}{\textit{Trained on MMSD2.0}}\\
SEAF-S & 71.65 & 64.73 & 65.47 & \textbf{69.82}\\
\method{} & \textbf{72.63} & \textbf{66.18} & \textbf{66.37} & 68.72\\
\bottomrule
\end{tabularx}
\caption{SarcNet transfer (\%): SEAF-S is semantic-only; $F_+$ is positive-class F1. Bold marks the best result within each source.}
\label{tab:transfer}
\end{table}

We use SEAF-S to test whether aligned visual evidence improves transfer beyond enriched semantic inputs: it retains post text, OCR, and entity background but removes the Aligned Visual Evidence Branch and SaCR loss.
We train both models separately on MMSD3.0 and MMSD2.0 and evaluate them on the complete official Chinese--English SarcNet test set \citep{yue2024sarcnet} without target training or tuning; source generation and encoding settings are retained.

\method{} achieves mean AUROCs of 84.29/72.63, exceeding SEAF-S by 4.64/0.98 points (Table~\ref{tab:transfer}).
These gains across both sources indicate that aligned visual evidence complements the same semantic inputs on unseen data, with the larger benefit in the primary MMSD3.0 setting.
Under the source-trained decision rule, macro-F1 changes by $-2.86$/$+1.45$; the consistent transfer benefit is therefore in threshold-independent ranking.
Appendix~\ref{app:literature} reports published zero-shot SarcNet results on the direct MLLM classification task.

\section{Input Selection and Architectural Analysis}
\label{sec:analysis}

\subsection{Full-Model Ablations}
\label{sec:ablation}

\begin{figure}[!htb]
\centering
\includegraphics[width=\columnwidth]{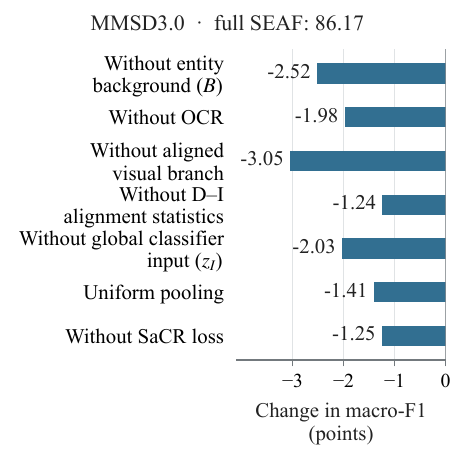}
\caption{MMSD3.0 ablation changes in macro-F1 (variant minus full SEAF; points). D--I: description--image alignment. Full results: Appendix~\ref{app:ablation}.}
\label{fig:ablation}
\end{figure}

To assess the contribution of each design choice, we compare full SEAF with variants under the same data splits and base training settings.
Each variant removes an evidence source or branch, zeros selected visual classifier inputs, replaces TASA with uniform pooling, or disables an auxiliary loss.
The resulting macro-F1 decreases indicate the complementary value of semantic and visual evidence, the benefit of text-conditioned aggregation, and the contribution of both contrastive objectives.
Figure~\ref{fig:ablation} summarizes MMSD3.0; Appendix~\ref{app:ablation} reports the complete results and intervention details.

\paragraph{Semantic and visual evidence.}
Removing the Aligned Visual Evidence Branch reduces macro-F1 by 3.05/6.28 points on MMSD3.0/MMSD2.0 while retaining post text, entity background, and OCR in the Semantic Evidence Branch, demonstrating the contribution of aligned visual evidence.
Removing entity background or OCR reduces macro-F1 by 2.52 or 1.98 points on MMSD3.0, respectively.
These inputs contribute entity context and image text to semantic interpretation.

\paragraph{Visual features and alignment.}
Setting the global visual classifier input $z_I$ to zero lowers MMSD3.0 macro-F1 by 2.03 points; setting the local input $p$ to zero lowers it by 0.99 points.
Excluding D--I alignment statistics from the visual classifier reduces macro-F1 by 1.24/1.13 points on MMSD3.0/MMSD2.0 with visual features retained.
The results show that both visual content and description-derived relations contribute to recognition, with descriptions supplying alignment evidence outside the semantic sequence.

\paragraph{Aggregation and regularization.}
Replacing TASA with uniform pooling reduces macro-F1 by 1.41 points on MMSD3.0, demonstrating the benefit of its text-conditioned weighting rule.
Disabling the SaCR loss or the semantic supervised contrastive loss reduces macro-F1 by 1.25 or 1.65 points, respectively.
These decreases show that both the within-post alignment constraint and the across-post semantic constraint contribute beyond branch-level cross-entropy supervision.

\subsection{Evidence Organization}
\label{sec:architecture_analysis}

\begin{figure}[!htb]
\centering
\includegraphics[width=\columnwidth]{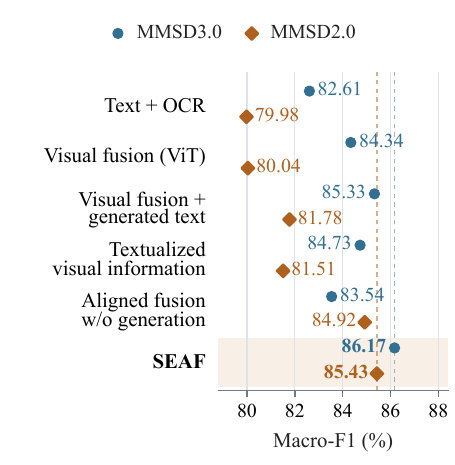}
\caption{Evidence organization: mean macro-F1 over three runs. Dashed lines mark SEAF; complete metrics appear in Table~\ref{tab:local_comparison}.}
\label{fig:evidence_organization}
\end{figure}

Figure~\ref{fig:evidence_organization} compares macro-F1 across alternative evidence organizations.
On MMSD3.0, \method{} exceeds visual fusion with generated text by 0.84 macro-F1 points and visual textualization by 1.44.
SEAF retains visual features and assigns background and descriptions to complementary roles.
On MMSD2.0, it exceeds aligned fusion without generation by 0.51 macro-F1 points.
These comparisons indicate the value of combining contextual supplementation with visual representations; the matched ablations assess individual components.

\subsection{Selecting Generated Content}
\label{sec:field_selection}

In an independent MMSD3.0 experiment, we vary only the generated field appended to the post text, keeping OCR, visual features, and D--I statistics fixed. This isolates the semantic value of the selected content.
Entity background yields 85.96 macro-F1, 2.13 points above complete neutral analyses and 2.26 above the variant without generated semantic input.
Full comparisons and length controls appear in Appendix~\ref{app:background}.

Additional controls show that complete analyses provide less benefit when aligned visual features are available (Appendix~\ref{app:explanation_interaction}).
Together with the ablation of D--I alignment statistics, these results favor the predefined allocation in Section~\ref{sec:neutral}: background supplements semantic context, descriptions provide alignment evidence, and image features retain visual content.

\section{Conclusion}

We introduce \method{}, which combines selective semantic enrichment with visual evidence fusion for multimodal sarcasm detection.
Entity background and OCR enrich semantic interpretation, while TASA aggregates global and local visual evidence using the original post text.
SEAF achieves the best overall results among compared methods on MMSD3.0; ablations demonstrate complementary contributions from its evidence sources and training objectives.

Our findings favor distinct roles for generated content: background supplements semantic context, while descriptions supply alignment evidence alongside retained visual features.
Complete analyses are less effective as semantic input than selected background; length-matched controls indicate that input brevity alone does not explain this advantage.
Enrichment should add missing context and image text while retaining complementary visual cues.

\FloatBarrier
\label{sec:content_end}

\section*{Limitations}

\paragraph{Evaluation scope.}
Our experiments focus on MMSD3.0 and MMSD2.0, with transfer to the bilingual SarcNet test set using models trained on English data.
Evaluation across additional languages, domains, and culturally specific references remains a direction for future work.

\paragraph{Generated context.}
SEAF uses neutral MLLM-generated background and descriptions as auxiliary evidence, whose usefulness depends on the relevance and accuracy of the generated content.
Our generator comparison examines two generation pipelines.
The effects of additional generators and varying availability of auxiliary metadata, such as product ratings, remain topics for further evaluation under consistent preprocessing.

\section*{Ethical Considerations}

Experiments use existing research datasets without new human annotation. Sarcasm depends on culture and context; predictions should not directly determine high-stakes judgments of user intent. 
Posts may contain offensive content or personal information. Released examples must follow licenses and be appropriately redacted. 
Generated background should not be redistributed as verified factual annotation.

Our planned release includes code and SEAF model weights. We verify that the applicable licenses and original creators' rights permit this release; source texts and images, OCR outputs, and generated analysis caches are not redistributed.

\begingroup
\microtypesetup{expansion=false}
\let\seafbibliography\thebibliography
\renewcommand{\thebibliography}[1]{%
  \seafbibliography{#1}%
  \tolerance=1000\relax
  \setlength{\emergencystretch}{5em}}
\bibliography{custom}
\endgroup
\clearpage
\appendix
\makeatletter
\setlength{\@fptop}{0pt}
\setlength{\@fpsep}{12pt}
\setlength{\@fpbot}{0pt plus 1fil}
\setlength{\@dblfptop}{0pt}
\setlength{\@dblfpsep}{12pt}
\setlength{\@dblfpbot}{0pt plus 1fil}
\makeatother

\section{Dataset Examples and Analysis Generation}
\label{app:motivation}

\subsection{Complementary Evidence in the Examples}

The examples in Figure~\ref{fig:motivation} illustrate the different information needed to interpret multimodal sarcasm.
The upper post invents a pardon for RFK to mock the denial of nepotism in Biden's pardon of his son.
Its background connects Biden's December 2024 pardon to RFK's 2014 Central Park bear incident and its public disclosure in August 2024.
The lower post pairs ``Rules of the Internet'' with a comic about fans disregarding a creator's boundaries.
OCR reveals the rules and dialogue that connect the images to the post's claim that younger generations no longer remember these rules.
The descriptions largely restate visible content; the background and OCR reveal the event context and specific wording needed to interpret the respective posts.

\subsection{Sensitivity to the Analysis Generator}
\label{app:generators}

Figure~\ref{fig:neutral_generation_prompt} gives the complete prompt, which requests literal descriptions and factual entity context while prohibiting sarcasm judgments and inferred author intent.
Table~\ref{tab:generator} tests whether the resulting auxiliary evidence can be used across generators, distinguishing replacement of test inputs from training with a different generator throughout.

\paragraph{Train on Qwen3-VL-8B inputs; vary the test generator.}
For each run, the upper block fixes the SEAF checkpoint trained and selected using Qwen3-VL-8B-generated background and descriptions.
At test time, the same classifier receives either Qwen3-VL-8B or Qwen3.8-flash inputs without retraining, reaching 86.17 and 86.09 macro-F1, respectively.
The 0.08-point difference indicates that these two generators provide compatible auxiliary evidence under the neutral prompting scheme.

\paragraph{Train and test with the same generator.}
The lower block trains a separate SEAF model for each generator, using its background and descriptions throughout training, validation, and testing.
Macro-F1 is 86.17 with Qwen3-VL-8B and 83.37 with Qwen3.8-flash.
This setting evaluates the complete generation-and-training pipeline, while the upper block isolates compatibility with a fixed classifier.
Both settings use the same neutral field types, allowing SEAF to obtain auxiliary evidence from either generator without asking it to predict sarcasm labels.

\begin{table*}[!t]
\centering\normalsize
\setlength{\tabcolsep}{8pt}
\renewcommand{\arraystretch}{1.0}
\begin{tabularx}{\textwidth}{@{}>{\raggedright\arraybackslash}X>{\raggedright\arraybackslash}Xrrrr@{}}
\toprule
\textbf{Train/dev input generator} & \textbf{Test input generator} & \textbf{Acc. (\%)} & \textbf{\boldmath$P_M$ (\%)} & \textbf{\boldmath$R_M$ (\%)} & \textbf{\boldmath$F_M$ (\%)}\\
\midrule
\multicolumn{6}{@{}l}{\textit{Fixed SEAF checkpoint per run; test-input generator varied}}\\
Qwen3-VL-8B & Qwen3-VL-8B & 86.76 & 86.00 & 86.36 & 86.17\\
Qwen3-VL-8B & Qwen3.8-flash & 86.64 & 85.82 & 86.43 & 86.09\\
\midrule
\multicolumn{6}{@{}l}{\textit{Separate SEAF models; same generator for training and testing}}\\
Qwen3-VL-8B & Qwen3-VL-8B & 86.76 & 86.00 & 86.36 & 86.17\\
Qwen3.8-flash & Qwen3.8-flash & 84.42 & 84.06 & 82.88 & 83.37\\
\bottomrule
\end{tabularx}
\caption{MMSD3.0 generator comparison (\%). For each run, the upper block fixes the SEAF checkpoint trained with Qwen3-VL-8B inputs and varies the test-input generator. 
The lower block trains and tests a separate SEAF model using each generator's inputs.}
\label{tab:generator}
\end{table*}

\section{Comparison Protocols and Reference Settings}
\label{app:literature}

\subsection{MMSD3.0 Input Protocols}

This comparison examines whether SEAF's advantage persists under a different adaptation of single-image baselines to multi-image posts.
For single-image baselines in the main comparison (Table~\ref{tab:main}), all images belonging to a post are tiled, in their original order, onto one canvas and then encoded as a single input.
In Table~\ref{tab:mmsd3_no_tiling}, each image is encoded separately with the same image encoder, and the resulting feature vectors are concatenated in image order.
The latter protocol preserves separate image encoding and combines information at the feature level.
These adaptations apply to single-image baselines; native multi-image methods retain their own input handling, including SEAF's per-image encoding and TASA aggregation (Section~\ref{sec:tasa}).
SEAF reports macro-averaged precision, recall, and F1; benchmark baselines retain their reported metric definitions.

For the Qwen3.7-Max zero-shot baseline, we use \texttt{qwen3.7-max-2026-06-08}, which supports image inputs \citep{alibabacloud2026qwen37max}.

SEAF reaches 86.76\% accuracy, higher than all alternatives in Table~\ref{tab:mmsd3_no_tiling}.
Its leading accuracy under both protocols shows that the comparison is not tied to canvas-based adaptation, complementing the controlled aggregation ablation in Section~\ref{sec:ablation}.

\begin{table*}[!t]
\centering\normalsize
\setlength{\tabcolsep}{4pt}
\renewcommand{\arraystretch}{1.0}
\begin{tabularx}{\textwidth}{@{}l>{\raggedright\arraybackslash}Xrrrr@{}}
\toprule
\textbf{Modality} & \textbf{Method} & \textbf{Acc. (\%)} & \textbf{P (\%)} & \textbf{R (\%)} & \textbf{F1 (\%)}\\
\midrule
\multirow{2}{*}{Image only}
& ResNet \cite{ResNet} & 65.33 & 53.13 & 36.18 & 43.05\\
& ViT \cite{dosovitskiy2021vit} & 64.03 & 50.31 & 55.12 & 52.61\\
\midrule
\multirow{5}{*}{Multimodal}
& DIP \cite{wen2023dip} & 81.49 & 81.17 & 79.95 & 80.41\\
& Multi-view CLIP \cite{qin2023mmsd2} & 82.20 & 75.86 & 79.81 & 77.79\\
& MoBA \cite{MoBA} & 80.17 & 79.26 & 78.83 & 79.02\\
& \citet{tang2024retrieval} & 82.39 & 81.47 & 81.61 & 81.54\\
& CIRM$^\dagger$ \cite{Zhao2026} & 83.44 & 82.65 & 82.44 & 82.54\\
\midrule
\multirow{1}{*}{Ours}
& \method{} & \textbf{86.76} & \textbf{86.00} & \textbf{86.36} & \textbf{86.17}\\

\bottomrule
\end{tabularx}
\caption{MMSD3.0 comparison with per-image encoding and feature concatenation for single-image baselines. Native multi-image methods retain their own input handling. Bold marks the best result for each metric. $\dagger$ marks reproduced results.}
\label{tab:mmsd3_no_tiling}
\end{table*}

\subsection{Zero-Shot SarcNet Reference}

Table~\ref{tab:sarcnet_vlm} lists all seven zero-shot MLLMs reported by \citet{basnet2025evalvlmsarcasm}, contextualizing direct MLLM classification on SarcNet.
This setting asks the MLLM itself to predict sarcasm.
The transfer experiment in Table~\ref{tab:transfer} instead tests whether SEAF's learned evidence fusion generalizes to SarcNet from a source dataset without target-domain training or tuning.
The two evaluations thus address different uses of pretrained models: direct classification and auxiliary evidence provision for a trained classifier.

\begin{table}[!t]
\centering\normalsize
\setlength{\tabcolsep}{5pt}
\renewcommand{\arraystretch}{1.0}
\begin{tabularx}{\columnwidth}{@{}>{\raggedright\arraybackslash}Xr@{}}
\toprule
\textbf{Model} & \textbf{Acc. (\%)}\\
\midrule
InstructBLIP & 67.0\\
Gemma3 & 66.0\\
PaliGemma & 57.0\\
LLaVA & 57.0\\
OpenFlamingo & 43.0\\
BLIP2 & 43.0\\
Qwen-VL & 43.0\\
\bottomrule
\end{tabularx}
\caption{Published zero-shot SarcNet accuracy (\%) from \citet{basnet2025evalvlmsarcasm}, using the source study's evaluation protocol.}
\label{tab:sarcnet_vlm}
\end{table}

\section{Generated Content Selection}
\label{app:background}

\subsection{Generated Fields and Fusion Strategy}
\label{app:field_replacement}

\begin{table*}[!t]
\centering\normalsize
\setlength{\tabcolsep}{8pt}
\renewcommand{\arraystretch}{1.0}
\begin{tabularx}{\textwidth}{@{}>{\raggedright\arraybackslash}Xlrrr@{}}
\toprule
\textbf{Generated semantic input} & \textbf{Adapter} & \textbf{Acc. (\%)} & \textbf{\boldmath$F_M$ (\%)} & \textbf{Change (pp)}\\
\midrule
Entity background (reference) & Off & 86.58 & 85.96 & --\\
No generated semantic input & Off & 84.11 & 83.70 & -2.26\\
Complete analysis & Off & 84.48 & 83.83 & -2.13\\
Analysis without entity background & Off & 84.24 & 83.45 & -2.51\\
Length-matched text summary & Off & 84.36 & 83.07 & -2.89\\
Entity background & On & 84.24 & 83.88 & -2.08\\
Complete analysis & On & 84.73 & 83.96 & -2.00\\
\bottomrule
\end{tabularx}
\caption{Independent MMSD3.0 field replacement: three-run means. Visual features and D--I statistics remain fixed; adapter-on variants add a separate fusion path and two auxiliary losses. 
Changes are in percentage points (pp), relative to this group\textquotesingle s background reference.}
\label{tab:selection}
\end{table*}

Table~\ref{tab:selection} examines which generated fields add useful semantic information beyond retained visual evidence, and whether a separate fusion path improves their use.
For field replacement, the adapter is off and only the generated content appended to the post text changes; visual features, D--I statistics, and the remaining settings stay fixed.
``Complete analysis'' includes entity background, image descriptions, and the other fields returned by the neutral prompt.
Changes are measured against this independent experiment's background reference of 85.96 macro-F1.

Removing entity background from the complete analysis tests whether its remaining fields can substitute for contextual knowledge.
This variant reaches 83.45 macro-F1, below both the complete analysis and selected background.
This result identifies entity context as a useful part of the generated output, while the stronger background-only result shows that appending the other fields does not necessarily improve its use.
The length-matched summary tests a different explanation for the gain, addressed in Appendix~\ref{app:input_length}.

\paragraph{Additional generated-text fusion.}
The adapter tests whether independently encoding the selected field adds value beyond encoding it jointly with the post text.
With the adapter \emph{off}, any selected field is appended to the post text, and no separate path is used.
With the adapter \emph{on}, this concatenation is retained, and the same field is independently encoded and projected into the fusion space.
Semantic representations after OCR cross-attention query this generated-text representation, which supplies keys and values.
The attention output enters a gated residual connection scaled by a learned example-level reliability score.
\paragraph{Adapter training objectives.}
Adapter-on variants add a reliability estimation loss $\mathcal L_{\mathrm{RQE}}$ and an auxiliary classification loss $\mathcal L_{\mathrm{aux}}$.
For training example $i$, let $h_i^-$ and $h_i^+$ denote the pooled representations before and after independent adapter injection.
The pre-injection representation still includes the selected field concatenated with the post text.
Two auxiliary classifiers, $f_-$ and $f_+$, compute losses against the sarcasm label $y_i$:
\begin{equation}
\begin{aligned}
e_i^-&=\operatorname{CE}(f_-(h_i^-),y_i),\\
e_i^+&=\operatorname{CE}(f_+(h_i^+),y_i).
\end{aligned}
\label{eq:adapter_aux_errors}
\end{equation}
The reliability predictor estimates the injection strength $q_i$ from $h_i^-$, with target and loss
\begin{equation}
\begin{aligned}
q_i^\star&=\operatorname{sg}\!\left[\sigma(e_i^- - e_i^+)\right],\\
\mathcal L_{\mathrm{RQE}}&=\frac{1}{n}\sum_{i=1}^{n}(q_i-q_i^\star)^2.
\end{aligned}
\label{eq:adapter_reliability}
\end{equation}
Here, $n$ is the batch size, $\sigma$ is the sigmoid function, and $\operatorname{sg}$ stops gradient propagation through the target.
A lower loss after injection relative to before injection yields a larger target strength.
Training labels provide this supervision without additional reliability annotations; at inference, $q_i$ is predicted from $h_i^-$ without labels.
The auxiliary classification loss trains both heads:
\begin{equation}
\mathcal L_{\mathrm{aux}}=\frac{1}{n}\sum_{i=1}^{n}(e_i^-+e_i^+).
\label{eq:adapter_auxiliary}
\end{equation}
The adapter-on controls add
\begin{equation}
\mathcal L_{\mathrm{extra}}=0.5\,\mathcal L_{\mathrm{RQE}}+0.3\,\mathcal L_{\mathrm{aux}}
\label{eq:adapter_extra}
\end{equation}
to the base training objective.
Final SEAF excludes the independent adapter and these extra losses, retaining the objective in Equation~\ref{eq:training-objective}.

Enabling the adapter changes macro-F1 from 85.96 to 83.88 for entity background and from 83.83 to 83.96 for complete analyses.
Neither variant improves on the background-concatenation reference.
These results favor SEAF's direct integration of selected background with the post text: the additional encoding and fusion path is not necessary to obtain the strongest result in this comparison.

\subsection{Text Length and Input Budget}
\label{app:input_length}

The length control tests whether entity background performs better merely because it produces shorter input.
Under the matched field-replacement setting in Table~\ref{tab:selection}, background reaches 85.96 macro-F1 versus 83.07 for a length-matched summary, a 2.89-point advantage.
Matching length therefore does not reproduce the benefit of supplying entity-specific context.

Table~\ref{tab:budget} separately examines how the generated fields use the 384-token total text budget.
Background and complete analyses have median lengths of 37 and 197 tokens, with over-budget rates of 0.0\% and 4.4\%, respectively.
Selected background thus supplies context more compactly and reduces exposure to truncation.
Together, the performance and length controls favor background for both its information content and its use of the available input budget.

\begin{table}[!t]
\centering\normalsize
\setlength{\tabcolsep}{5pt}
\renewcommand{\arraystretch}{1.0}
\begin{tabularx}{\columnwidth}{@{}>{\raggedright\arraybackslash}Xrr@{}}
\toprule
\textbf{Input} & \shortstack[r]{\textbf{Median}\\\textbf{tokens}} & \shortstack[r]{\textbf{Over budget}\\\textbf{(\%)}}\\
\midrule
Entity background & 37 & 0.0\\
Complete analysis & 197 & 4.4\\
\bottomrule
\end{tabularx}
\caption{MMSD3.0 generated-input lengths and truncation, with total text budget 384 tokens.}
\label{tab:budget}

\end{table}

\subsection{Complete Analyses with and without Visual Evidence}
\label{app:explanation_interaction}

These controls test whether the benefit of a complete neutral MLLM-generated analysis $R$ depends on the visual evidence already available to the classifier.
The independent text control $\tcore$ uses post text and OCR; $\vcore$ adds an aligned visual stream.
Before adding $R$, both omit entity background, D--I statistics, and SaCR.
For each control, we compare classification with and without appending $R$ to the semantic sequence, keeping its other inputs and settings unchanged.
This design measures the incremental value of the same generated content in two evidence settings.

\begin{figure}[!htb]
\centering
\includegraphics[width=\columnwidth]{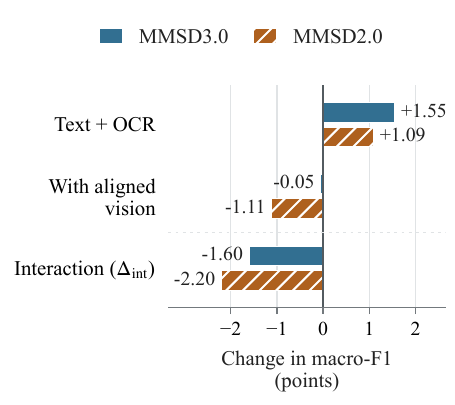}
\caption{Effect of adding a complete neutral generated analysis $R$ to the semantic input. Bars show changes in macro-F1 (points); $\Delta_{\mathrm{int}}=\Delta_V-\Delta_T$ compares the benefit with and without aligned vision.}
\label{fig:analysis_interaction}
\end{figure}

Figure~\ref{fig:analysis_interaction} reports the effect of adding $R$ to the text control ($\Delta_T$) and to the control with aligned vision ($\Delta_V$).
On MMSD3.0/MMSD2.0, the text-control gains are 1.55/1.09 points, whereas the aligned-visual changes are $-0.05$/$-1.11$ points.
The difference $\Delta_{\mathrm{int}}=\Delta_V-\Delta_T$ is $-1.60$/$-2.20$ points.
Complete analyses help the text control but add no improvement once aligned visual features are available.
This pattern is consistent with overlap between descriptive text and visual features: the useful semantic supplement should depend on which evidence the model already retains.
Together with the field-replacement results, it favors adding targeted entity context to a model that preserves visual evidence.

\section{Complete Ablations and Intervention Details}
\label{app:ablation}

\subsection{Cross-Dataset Component Effects}

Table~\ref{tab:ablation_complete} extends Section~\ref{sec:ablation} with single-image results and the complete set of component controls.
It tests whether the evidence sources and training objectives remain useful across input settings; the intervention definitions below specify exactly what is removed or replaced.

\begin{table*}[!t]
\centering\normalsize
\setlength{\tabcolsep}{8pt}
\renewcommand{\arraystretch}{1.0}
\begin{tabularx}{\textwidth}{@{}>{\raggedright\arraybackslash}Xrrrr@{}}
\toprule
\textbf{Model variant} & \multicolumn{2}{c}{\textbf{MMSD3.0}} & \multicolumn{2}{c}{\textbf{MMSD2.0}}\\
\cmidrule(lr){2-3}\cmidrule(lr){4-5}
 & \textbf{\boldmath$F_M$ (\%)} & \textbf{Change (pp)} & \textbf{\boldmath$F_M$ (\%)} & \textbf{Change (pp)}\\
\midrule
SEAF (full model) & 86.17 & -- & 85.43 & --\\
\midrule
Without entity background ($B$) & 83.65 & -2.52 & 84.22 & -1.21\\
Without OCR & 84.19 & -1.98 & 84.06 & -1.37\\
Without aligned visual branch & 83.12 & -3.05 & 79.15 & -6.28\\
Without semantic branch & 61.85 & -24.32 & -- & --\\
Without D--I alignment statistics & 84.93 & -1.24 & 84.30 & -1.13\\
Without alignment statistics ($u$) & 84.98 & -1.19 & -- & --\\
Without global classifier input ($z_I$) & 84.14 & -2.03 & -- & --\\
Without local classifier input ($p$) & 85.18 & -0.99 & -- & --\\
Uniform pooling & 84.76 & -1.41 & -- & --\\
Without SaCR loss & 84.92 & -1.25 & 84.30 & -1.13\\
Without semantic supervised contrastive loss & 84.52 & -1.65 & -- & --\\
\bottomrule
\end{tabularx}
\caption{Complete component ablations: three-run mean macro-F1 and change from full SEAF (variant minus full; pp). Negative values indicate lower macro-F1. The references are 86.17 on MMSD3.0 and 85.43 on MMSD2.0. Feature and statistic ablations set the corresponding visual classifier inputs to zero. Dashes denote unavailable controls.}
\label{tab:ablation_complete}
\end{table*}

On MMSD2.0, removing entity background, OCR, or SaCR reduces macro-F1 by 1.21, 1.37, or 1.13 points, respectively.
Their contributions therefore extend to single-image inputs, where there is no need to weight multiple images.
This separates the value of contextual supplementation and alignment regularization from the benefit of multi-image aggregation.

On MMSD3.0, excluding all alignment statistics reduces macro-F1 by 1.19 points, while excluding only D--I statistics reduces it by 1.24.
Both comparisons show the value of relational evidence alongside visual features.
The decreases measure contributions conditional on the remaining inputs, so they do not form an additive decomposition of the full model's performance.

\subsection{Intervention Scope}

\paragraph{Input ablations.}
These controls test the contribution of auxiliary context while preserving original post text and images.
\emph{Without entity background ($B$)} omits $B$ from the semantic sequence while retaining descriptions for D--I alignment.
\emph{Without OCR} removes OCR from semantic cross-attention and sets the O--I alignment statistics to zero.

\paragraph{Branch ablations.}
These controls assess the complementary roles of contextual semantic modeling and aligned visual evidence.
\emph{Without aligned visual branch} is the SEAF-S variant: it retains post text, OCR, and entity background in the Semantic Evidence Branch while removing the visual classifier, alignment statistics, and SaCR loss.
\emph{Without semantic branch} removes semantic encoding, classification, and semantic objectives, retaining the Aligned Visual Evidence Branch with its aligned text features and OCR/description statistics.
Its 61.85 macro-F1 on MMSD3.0, compared with 86.17 for full SEAF, indicates the importance of contextual semantic modeling in the complete framework.
Trainable parameters change from 33.88M to 1.58M, so this comparison assesses the full semantic path, including its capacity and training objectives.

\paragraph{Classifier-input ablations.}
These controls distinguish the information carried by global features, local features, and alignment statistics at the visual classifier.
\emph{Without global classifier input ($z_I$)} sets $z_I$ to zero only at that input, retaining the global projection for SaCR.
\emph{Without local classifier input ($p$)} zeros the pooled local input $p$.
\emph{Without D--I alignment statistics} zeros the description--image mean, minimum, and maximum coordinates in $u$; \emph{Without alignment statistics ($u$)} zeros the entire vector.
Each variant retains the other classifier inputs, so removing descriptive relations does not remove visual features.

\paragraph{Aggregation and objective ablations.}
\emph{Uniform pooling} replaces TASA with equal image weights ($w_k=1/K$) on both global and local pooling paths, leaving alignment statistics unchanged.
It tests TASA's complete similarity-based, smoothed weighting rule against equal aggregation.
The objective controls retain the architecture and inputs to distinguish the effect of training supervision from a structural change.
\emph{Without SaCR loss} and \emph{Without semantic supervised contrastive loss} disable only $\mathcal L_{\mathrm{SaCR}}$ and $\mathcal L_{\mathrm{SupCon,S}}$, respectively, in Equation~\ref{eq:training-objective}.

\section{Additional Architecture Results}
\label{app:visual}

\subsection{Complete Evidence-Organization Results}

\begin{table*}[!t]
\centering\normalsize
\setlength{\tabcolsep}{4pt}
\renewcommand{\arraystretch}{1.0}
\begin{tabularx}{\textwidth}{@{}>{\raggedright\arraybackslash}Xrrrrrrrr@{}}
\toprule
\multirow{2}{*}{\textbf{Model}} & \multicolumn{4}{c}{\textbf{MMSD3.0}} & \multicolumn{4}{c}{\textbf{MMSD2.0}}\\
\cmidrule(lr){2-5}\cmidrule(lr){6-9}
 & \shortstack[r]{\textbf{Acc.}\\\textbf{(\%)}} & \shortstack[r]{\textbf{\boldmath$P_M$}\\\textbf{(\%)}} & \shortstack[r]{\textbf{\boldmath$R_M$}\\\textbf{(\%)}} & \shortstack[r]{\textbf{\boldmath$F_M$}\\\textbf{(\%)}} & \shortstack[r]{\textbf{Acc.}\\\textbf{(\%)}} & \shortstack[r]{\textbf{\boldmath$P_M$}\\\textbf{(\%)}} & \shortstack[r]{\textbf{\boldmath$R_M$}\\\textbf{(\%)}} & \shortstack[r]{\textbf{\boldmath$F_M$}\\\textbf{(\%)}}\\
\midrule
Text + OCR & 83.50 & 82.71 & 82.52 & 82.61 & 80.03 & 80.29 & 80.86 & 79.98\\
Visual fusion (ViT) & 85.34 & 85.10 & 83.81 & 84.34 & 80.07 & 80.62 & 81.11 & 80.04\\
Visual fusion + generated text & 86.21 & 85.87 & 84.92 & 85.33 & 81.86 & 81.93 & 82.56 & 81.78\\
Textualized visual information & 85.53 & 84.90 & 84.59 & 84.73 & 81.53 & 82.18 & 82.65 & 81.51\\
Aligned fusion w/o generation & 83.93 & 83.21 & 84.61 & 83.54 & 85.01 & 84.91 & \textbf{85.58} & 84.92\\
\method{} & \textbf{86.76} & \textbf{86.00} & \textbf{86.36} & \textbf{86.17} & \textbf{85.68} & \textbf{85.35} & 85.53 & \textbf{85.43}\\
\bottomrule
\end{tabularx}
\caption{Evidence-organization controls: P/R/F1 are macro averages; bold indicates the best result for each metric within each dataset.}
\label{tab:local_comparison}
\end{table*}

Table~\ref{tab:local_comparison} complements Figure~\ref{fig:evidence_organization} by comparing how different architectures represent and combine the available evidence.
The text-and-OCR control provides a textual reference; the ViT fusion variants examine visual features with and without generated text; visual textualization provides a text-based representation of visual content; and aligned fusion without generation evaluates aligned evidence without generated context.
SEAF combines retained visual features with the selected background and description-derived relations.
These are comparisons of complete evidence organizations, while Appendix~\ref{app:ablation} isolates designated components within SEAF.

SEAF's leading macro-F1 is accompanied by the best accuracy and macro-precision on both datasets.
The agreement across these metrics shows that its advantage extends beyond a single summary score and favors combining selective contextual enrichment with retained visual evidence.

\subsection{Independent Visual Controls}

\begin{table*}[!t]
\begin{minipage}[t]{0.48\textwidth}\vspace{0pt}
\centering\normalsize
\setlength{\tabcolsep}{5pt}
\renewcommand{\arraystretch}{1.0}
\begin{tabularx}{\linewidth}{@{}>{\raggedright\arraybackslash}Xrr@{}}
\toprule
\textbf{Change} & \textbf{MMSD3.0} & \textbf{MMSD2.0}\\
\midrule
Add ViT to text control & +1.73 & +0.06\\
Add aligned vision to $\tcore$ & +1.07 & +4.71\\
Aligned vs. ViT fusion & -0.80 & +4.88\\
\bottomrule
\end{tabularx}
\caption{Independent visual-control differences of three-run mean macro-F1 in percentage points (pp). Each path uses its own text control.}
\label{tab:visual}
\end{minipage}\hfill
\begin{minipage}[t]{0.48\textwidth}\vspace{0pt}
\centering\normalsize
\setlength{\tabcolsep}{5pt}
\renewcommand{\arraystretch}{1.0}
\begin{tabularx}{\linewidth}{@{}>{\raggedright\arraybackslash}Xrr@{}}
\toprule
\textbf{Model} & \textbf{Params. (M)} & \textbf{\boldmath$F_M$ (\%)}\\
\midrule
ViT visual fusion & 64.20 & 84.34\\
\method{} & \textbf{33.88} & \textbf{86.17}\\
\bottomrule
\end{tabularx}
\caption{Trainable task parameters (millions) and MMSD3.0 mean macro-F1, excluding the frozen encoders and generator.}
\label{tab:capacity}
\end{minipage}
\end{table*}

These controls examine whether aligned visual representations contribute useful information even before generated evidence is introduced.
We reuse the independent $\tcore$/$\vcore$ pair defined in Appendix~\ref{app:explanation_interaction}, alongside the text and ViT fusion pipeline.
Each visual gain is calculated relative to its own text control, keeping the two base pipelines distinct.
Table~\ref{tab:visual} shows that adding aligned vision improves macro-F1 by 1.07/4.71 points on MMSD3.0/MMSD2.0.
The positive gains on both datasets establish the value of retained visual evidence without attributing that gain to generated background or descriptions.

\section{Metric Definitions and Resource Accounting}
\label{app:evaluation}

\subsection{Metric Definitions}

Macro-precision, macro-recall, and macro-F1 average the corresponding metrics over the sarcastic and non-sarcastic classes. 
Macro-F1 averages class-wise F1 scores, rather than taking the harmonic mean of macro-precision and macro-recall. Zero-denominator metrics are set to zero. 
Positive-class F1 uses the sarcastic class. AUROC ranks sarcasm scores independently of the fixed classification decision, assigning half credit to tied positive--negative pairs.

\subsection{Resource Scope}
\label{app:resources}

Table~\ref{tab:capacity} examines the trainable model size associated with the predictive gains.
SEAF uses 33.88M trainable task parameters versus 64.20M for ViT fusion, a 47.2\% reduction, while improving MMSD3.0 macro-F1 from 84.34 to 86.17.
The improvement is therefore achieved with a smaller trainable task model in this comparison.

The counts exclude the generator and frozen encoders.
Caching generated inputs and frozen features avoids repeating this preprocessing across training epochs; unseen posts still require generation, OCR, and feature extraction.
These counts measure trainable model size, while end-to-end latency, memory, and energy additionally depend on preprocessing, hardware, and workload.

\section{Implementation and Training Details}
\label{app:implementation}

\paragraph{Encoders and cached inputs.}
We use DeBERTaV3 \citep{he2023debertav3} as the text encoder and freeze its first eight layers. CLIP ViT-L/14@336px \citep{radford2021clip} serves as the frozen shared-space encoder. Qwen3-VL-8B generates neutral auxiliary inputs subject to the constraints in Section~\ref{sec:neutral}. Generated image descriptions, entity background information, and features extracted by frozen encoders are cached for the corresponding training and evaluation examples and reused across epochs.

\paragraph{Model and input settings.}
The hidden dimension is 512, with eight attention heads and a $7\times7$ local feature grid. The maximum sequence lengths are 384 tokens for semantic text, 256 for OCR text, and 77 for aligned text. Entity background information is additionally capped at 2,000 characters and included within the semantic-text token budget. We set $\alpha=0.3$, $\tau=0.05$, $m=0.2$, and $\lambda_{\mathrm{SC}}=0.2$.

\paragraph{Optimization and checkpoint selection.}
We train for 12 epochs with a batch size of 8 and two gradient accumulation steps. The learning rates for the encoder and fusion modules are $2\times10^{-5}$ and $2\times10^{-4}$, respectively. We use a weight decay of 0.01, a warmup ratio of 0.1, a dropout rate of 0.3, a gradient clipping threshold of 0.3, and a label smoothing factor of 0.1. For each run, we select the checkpoint with the highest positive-class F1 on the validation set; the primary test metric is macro-F1.

\paragraph{Run aggregation.}
\method{} results, component ablations, and transfer results are means over three runs. 
Component changes are calculated within each matched experimental group, using its own full-model reference. 
Differences in this paper are calculated from the reported two-decimal means.

\clearpage
\begin{figure*}[p]
\centering
\begin{minipage}{0.95\textwidth}
\begin{promptpanel}{System Prompt for Neutral Analysis}
\begin{Verbatim}[fontsize=\small,commandchars=\\\^\~,breaklines=true,breakanywhere=true,breakautoindent=true,breakindent=0pt,breaksymbolleft={},breaksymbolright={},tabsize=2]
You are an objective multimodal post analyzer. Given a Twitter or Amazon post with one or more images and text, produce a NEUTRAL, FACT-BASED structured description in JSON.

CRITICAL RULES:
- NEVER use the words: sarcastic, sarcasm, ironic, irony, mocking, \seafpromptterma^~, \seafprompttermb^~, \seafprompttermc^~, \seafprompttermd^~, \seafpromptterme^~, or any synonyms / translations / paraphrases of these.
- Do NOT judge whether the post is sarcastic or sincere, and do NOT state the author's true intent or hidden meaning.
- Describe only what is LITERALLY SHOWN or LITERALLY SAID.
- The text_summary field must report what the words on the page mean by surface reading, even if it sounds inconsistent with the rating or images.
- In the incongruity_analysis field ONLY, you MAY point out OBSERVABLE factual MISMATCHES between the literal text, the images, and the star rating (e.g. "the text praises the product but the image shows it broken"). State these strictly as neutral observations of factual inconsistency — never label them, never use the forbidden words, never conclude what the author really means or feels.

Output STRICT JSON. No prose before or after. No markdown fences.
\end{Verbatim}
\end{promptpanel}
\vspace{3pt}
\begin{promptpanel}{User Prompt and JSON Output Format}
\begin{Verbatim}[fontsize=\small,commandchars=\\\^\~,breaklines=true,breakanywhere=true,breakautoindent=true,breakindent=0pt,breaksymbolleft={},breaksymbolright={},tabsize=2]
Source: {source}
Star rating: {star}
Post text: {text}

Analyze the {n_images} image(s) above and respond in this exact JSON shape (no extra fields, no comments):

{
  "per_image": [
    {"index": 1, "description": "<1-2 sentences: visible objects, scene, on-image text>"}
  ],
  "cross_image_relation": {
    "type": "<one of: contrast | complement | sequential | unrelated>",
    "detail": "<1 sentence>"
  },
  "entities": [
    {"name": "<named entity>", "type": "<person|brand|product|location|meme|cultural_ref|other>", "background": "<1 sentence factual context>"}
  ],
  "text_summary": "<1 sentence: surface reading of the text>",
  "incongruity_analysis": "<1-2 sentences: observable factual mismatch among the literal text, the images, and the star rating; write 'none' if text, images and rating are mutually consistent>"
}

Rules:
- per_image: one entry per image in order; index starts from 1.
- cross_image_relation: use "unrelated" if a single image or independent images.
- entities: list named people/brands/products/cultural references a general reader may not know; empty list if none.
- text_summary: literal surface meaning only. Do not contrast with rating. Do not use forbidden words.
- incongruity_analysis: report ONLY observable inconsistencies (text vs image vs rating) as neutral fact; do NOT label, do NOT state intent, do NOT use forbidden words; write "none" if consistent.
\end{Verbatim}
\end{promptpanel}
\end{minipage}
\caption{System instructions and user template for neutral multimodal analysis.}
\label{fig:neutral_generation_prompt}
\end{figure*}

\end{document}